# Natural Scene Character Recognition Using Robust PCA and Sparse Representation

Zheng Zhang, Yong Xu
Bio-Computing Research Center,
Harbin Institute of Technology, Shenzhen Graduate School,
Shenzhen 518055, P.R. China
darrenzz219@gmail.com; yongxu@ymail.com

Cheng-Lin Liu
National Laboratory of Pattern Recognition, CEBSIT,
Institute of Automation, Chinese Academy of Sciences
Beijing 100190, P.R. China
liucl@nlpr.ia.ac.cn

*Abstract*—Natural scene character recognition is challenging due to the cluttered background, which is hard to separate from text. In this paper, we propose a novel method for robust scene character recognition. Specifically, we first use robust principal component analysis (PCA) to denoise character image by recovering the missing low-rank component and filtering out the sparse noise term, and then use a simple Histogram of oriented Gradient (HOG) to perform image feature extraction, and finally, use a sparse representation based classifier for recognition. In experiments on four public datasets, namely the Char74K dataset, ICADAR 2003 robust reading dataset, Street View Text (SVT) dataset and IIIT5K-word dataset, our method was demonstrated to be competitive with the state-of-the-art methods.

## I. Introduction

Scene text recognition has attracted increasing attention [1] in recent years due to its significance in scene understanding and widespread applications in intelligent systems. Automatically understanding the text information in images is in urgent need for many applications such as robot navigation, image understanding and retrieval, unmanned scene recognition and sign recognition [2].

Over the past several decades, extensive optical character recognition (OCR) methods [3][4] have been developed for extracting text information from images. These conventional OCR methods perform satisfactorily on scanned document images, but for extracting text from natural scene images, they cannot work well because the separation of text from image background is very hard. Unlike scanned paper documents, scene text images are usually captured under uncontrolled environments. Therefore, scene text image recognition suffers from many obstacles and difficulties, such as illumination variations, complex background, low-resolution, occlusion, blurring, the changes of text size, font, color, line orientation and position [5]. Fig. 1 shows some cropped scene character images, which manifest a multitude of style variations.

Extensive scene text recognition methods have been proposed to achieve better recognition results [6][7]. Some scene text recognition methods were still based on existing OCR engines after image preprocessing such as image binarization and slant correction. For example, Chen and Yuille [8] proposed to first binarize the text image using an improved adaptive binarization algorithm, and then a commercial OCR system was used for recognition. Wakahara and Kita [9] devised a method of binarizing multicolored scene character strings using iterative K-means clustering and support vector machines. However, binarization techniques cannot segment the texts from scene images perfectly, largely due to diversity and complexity of image variations. The PhotoOCR of Google [10] uses conventional OCR on multiple binarized images to overcome the insufficiency of binarization.

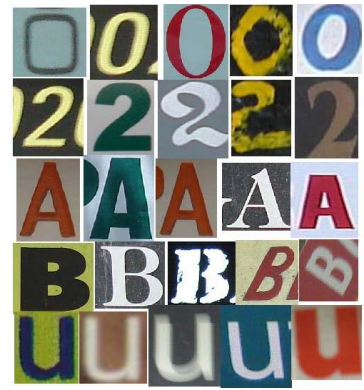

Fig. 1. Example scene character images selected from the Char74k goodImg dataset. Each row shows the samples of one class.

Recently, some researchers turn to develop methods for scene text recognition on color or gray-scaled images without binarization, and have achieved results superior to OCR engines [11]. Specifically, the recently proposed methods usually extract informative features from the gray or color images to overcome the cluttered background. This class of method was compared with OCR engines by combining two classifiers with six feature vectors [12]. The features there include different features representing texture and different types of shape and edge features such as Shape Context, Geometric Blur (GB) and SIFT. The classifiers include the Nearest Neighbor (NN) classifier and SVM with different kernel functions.

In this paper, we propose a method for robust scene character recognition using robust principal component analysis (PCA) [13] and sparse representation based classification (SRC) [14]. Extensive experimental results on different datasets show that the proposed method can achieve competent performance in scene character recognition.

The remainder of this paper is organized as follows. Section II briefly reviews the related work. Section III presents the proposed robust scene character recognition method. Experiments on scene character datasets are discussed in Section IV, and Section V gives concluding remarks.

## II. RELATED WORKS

The success of robust PCA has attracted increasing research attention for recovering clean images and eliminating noises. For example, robust PCA [13] not only can recover missing components but also remove shadows and specularities from face images under different illuminations. The GoDec algorithm [15] exploits the bilateral random projections to efficiently decompose a matrix into a low-rank part and a sparse noise part. Subsequently, more robust low-rank representation based methods have been proposed to perform image denoising [16].

The conventional HOG feature has been widely used for scene character recognition. As demonstrated in [2][17][18], the recognition results based on gradient feature in most cases outperform all the other local sampling based feature extraction methods. The pyramid of HOG feature is integrated with chi-square kernel based SVM to perform scene character recognition and the result is comparable to the current techniques [19]. Yi et al. [20] verified the effectiveness of the gradient feature by evaluating different feature representations for scene character recognition, and developed a global HOG method, which directly extracts HOG feature from the whole character image. Recently, many modified HOG features have been developed to improve scene character recognition, such as Co-HOG [6] and ConvCoHOG [11]. Moreover, a method based on histograms of sparse coding features [21] is proposed for scene text recognition and obtains competent recognition results, while we use the sparse codes for classification.

Sparse representation has been proven to be a powerful technique to a wide range of applications, especially in signal processing, image processing, pattern recognition and computer vision [14][22]. A recent survey [22] comprehensively reviewed the most representative sparse representation algorithms and explicitly summarized its main applications. Sparse representation based classification (SRC) method has been demonstrated as a robust and effective method in image classification [14] [22].

## III. METHOD DESCRIPTION

The flowchart of our method is presented in Fig. 2. We first exploit the robust PCA to recover clean character images from the blurred or corrupted images, and then use the HOG method to extract gradient features from the recovered image, and utilize the SRC method to perform recognition. In this section, we first give a explicit description of robust PCA [13] and SRC [14], and then elaborate on the details of our method.

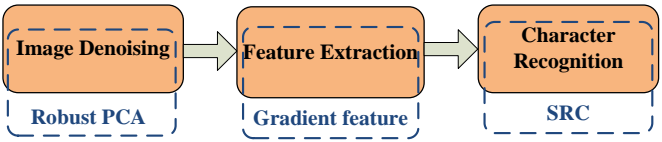

Fig. 2. Flowchart of our proposed method.

Algorithm 1. Robust PCA via the inexact ALM [23]

Input: Observed data matrix $X$, $\lambda > 0$, $\rho = 1.5$, $k = 0$

While not converged do

Step 1: Exploiting SVT algorithm to solve problem (4):

$[U\ S\ V] = svd(X - E^k + Y^k/\mu^k)$, $X_0^{k+1} = U\,soft_{1/\mu^k}(S)V^T$;

Step 2: $E^{k+1} = soft_{\lambda/\mu}(X - X_0^{k+1} + Y^k/\mu^k)$;

Step 3: $Y^{k+1} = Y^k + \mu^k(X - X_0^{k+1} - E^{k+1})$;

Step 4: $\mu^{k+1} = \min(\rho\mu^k, 10^5)$;

End

Output: $X_0^{k+1}, E^{k+1}$

### A. Robust principle component analysis

Let $X = [x_1, x_2, ..., x_n] \in \mathbb{R}^{d \times n}$ be the observed data matrix, which consists of $n$ samples, and each column is an image vector. Considering that the observed data such as scene character images usually contain noise, recovering clean data is an essential procedure before recognition. The goal of the RPCA method [13] is to recover a clean low-rank matrix $X_0$ from the corrupted data matrix, i.e. $X = X_0 + E$. Moreover, $E$ usually is a sparse matrix because noise is only a small component of each image. Thus, the problem of RPCA can be formulated as the following regularized rank minimization problem:

$$\arg\min_{X_0, E} rank(X_0) + \lambda \|E\|_0 \ \ s.t.\ X = X_0 + E. \quad (1)$$

where $\lambda$ is the Lagrange multiplier for balancing the low-rank term and the noise term, and $\|\bullet\|_0$ is the $l_0$ pseudo-norm, i.e. the number of the nonzero entries in a matrix or vector.

Because of the discrete properties of the rank function and the $l_0$ norm, it is difficult to solve the above minimization problem (1), which is a non-convex problem. We reformulate problem (1) into a tractable optimization problem by respectively replacing the rank function with the nuclear norm and the $l_0$ norm with the $l_1$ norm [13], i.e.

$$\arg\min_{X_0, E} \|X_0\|_* + \lambda \|E\|_1 \ \ s.t.\ X = X_0 + E. \quad (2)$$

where $\|\bullet\|_*$ is the nuclear norm, i.e. the sum of the singular values, and $\|\bullet\|_1$ is the $l_1$ norm, i.e. the sum of the absolute values of the matrix entries. The optimization of problem (2) can be solved by the Augmented Lagrange Multiplier (ALM) method [23] and the augmented Lagrangian function of (2) is reformulated as

$$\min_{X_0, E} \|X_0\|_* + \lambda \|E\|_1 + <Y, X - X_0 - E> + \frac{\mu}{2}\|X - X_0 - E\|_F^2. \quad (3)$$

which can be rewritten as

$$\min_{X_0, E} \|X_0\|_* + \lambda \|E\|_1 + \frac{\mu}{2}\|X - X_0 - E + \frac{Y}{\mu}\|_F^2. \quad (4)$$

where $\mu$ is a positive scalar and $Y$ is a vector of Lagrange multipliers. Thus, the optimization of problem (4) can be solved by alternatively updating $X_0$ and $E$ when fixing others. More specifically, when $E$ is given, problem (4) is converted to the following equivalent optimization problem

$$\min_{X_0} \| X_0 \|_* + \frac{\mu}{2} \| X_0 - (X - E + \frac{Y}{\mu}) \|_F^2 . \qquad (5)$$

When $X_0$ is fixed, problem (4) is reduced to the following problem:

$$\min_{E} \lambda \| E \|_1 + \frac{\mu}{2} \| E - (X - X_0 + \frac{Y}{\mu}) \|_F^2 . \qquad (6)$$

For convenient description of optimizing (4), we introduce a simple well-know soft-thresholding or shrinkage operator [22] as follows:

$$soft_\lambda(a) = \begin{cases} a - \lambda, & \text{if } a > \lambda \\ a + \lambda, & \text{if } a < -\lambda \\ 0, & \text{otherwise.} \end{cases} \qquad (7)$$

Thus, the following optimization problem

$$\arg\min_X \lambda \| X \|_1 + \frac{1}{2} \| X - A \|_F^2 . \qquad (8)$$

can be easily solved by $X = soft_\lambda(A)$ [19], and the following nuclear norm minimization

$$\arg\min_X \lambda \| X \|_* + \frac{1}{2} \| X - A \|_F^2 . \qquad (9)$$

can be solved by Singular Value Thresholding (SVT) method.

The optimization of problem (4) can be efficiently solved by inexact ALM presented in [23] and all the procedures are summarized in Algorithm 1.

*B. Sparse representation based classification method*

Suppose that there are $m$ training samples from $c$ classes, which are stacked to a matrix $D = [d_1, d_2, ..., d_m]$. Let $D_i$ denote the samples from the $i$-th class and the testing sample is a vector $y$. SRC [14, 22] assumes that given sufficient training samples, each test sample can be approximately represented by the linear combination of just those training samples from the same class. SRC exploits all the training samples to represent the testing sample and the following equation is satisfied

$$\arg\min_z \| z \|_0 \ s.t. \ y = Dz . \qquad (10)$$

where $z = [z_1, z_2, ..., z_m]$. Here $z$ is a sparse vector whose most entries are zero except those associated with the same class of $y$. The theory of compressed sensing has demonstrated that if the solution of $z$ is sufficiently sparse [14], problem (10) can be equivalently solved by the following $l_1$ norm minimization problem:

$$\arg\min_z \| z \|_1 \ s.t. \ y = Dz . \qquad (11)$$

Because real data always contains noise, representation noise is unavoidable in the process of sparse representation. If the representation noise is bounded to a small constant $\varepsilon$, problem (11) can be rewritten as

$$\arg\min_z \| z \|_1 \ s.t. \| y - Dz \|_2^2 \leq \varepsilon . \qquad (12)$$

and its unconstrained optimization problem is reformulated as

$$\arg\min_z \lambda \| z \|_1 + \| y - Dz \|_2^2 \qquad (13)$$

where $\lambda$ refers to the Lagrange multiplier associated with $\| z \|_1$. We exploit the Homotopy based sparse representation method to solve problem (13) because of its efficiency and high accuracy [22]. The main scheme of the SRC method has been summarized in Algorithm 2.

---

Algorithm 2. A general framework of SRC method [22]

---

Input: Training sample matrix $D$, $\lambda > 0$

Step 1: Normalize all the samples to have unit vectors;

Step 2: Represent the test sample $y$ over $D$ exploiting the $l_1$ norm minimization problem (13);

Step 3: Compute the representation residual of each class $r_i = \| y - D_i z_i \|_2^2$ where $z_i$ denotes the representation coefficient vector associated with class $i$;

Step 4: Output the label of the test sample $y$ using $identity(y) = \arg\min_i(r_i)$.

---

*C. The proposed method for scene character recognition*

Our scene character recognition method consists of three steps, image denoising using robust PCA, extraction of HOG features and character recognition using SRC.

The motivations of our proposed method are as follows. Sparse representation has become a fundamental technique in different fields and can achieve satisfactory results in image classification, but has not been applied into scene character recognition. Recently proposed robust PCA can effectively eliminate noises in the images and integrating robust PCA and SRC is a reasonable way to obtain better recognition results. On the other hand, considering the particular characteristics of scene character images, the performance of extracting gradient feature is very impressive and a clear strategy is formulated as: image denoising based on robust PCA, feature extraction using a simple HOG and character recognition using SRC. The detailed procedures are summarized as follows.

First, we use robust PCA (presented in Algorithm 1) to denoise scene character images. As we know, if the training samples are heavily corrupted, the performance of any classifier will degenerate. Fortunately, robust PCA can remove the noise term by decomposing the corrupted image into a clean image term and sparse noise term. By removing the sparse noise term, images from the same class can have more similar features. Some denoising results using robust PCA are shown in Fig. 3. We can see that some irrelevant noise terms have been removed and more useful information are simultaneously recovered. For example, as illustrated from the first row of Fig. 3, the blurred number image '0' has become more clear and some more details has been restored. Moreover, from columns (3) and (6) in Fig. 3, we can see that some sparse noises have been filtered out.

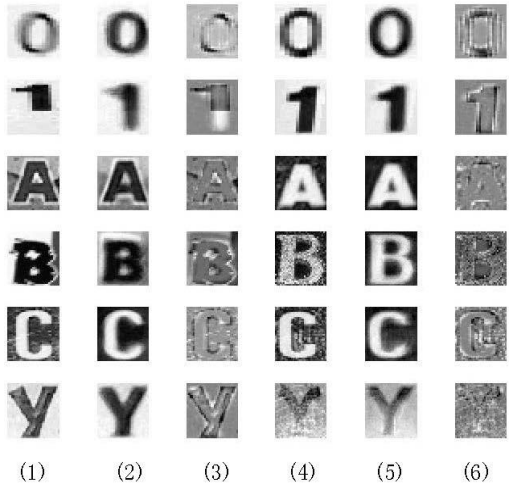

Fig. 3. Some image denoising results obtained by using robust PCA on some samples from the ICDAR 2003 datasets. Colomns (1) and (4) are the original gray scale images, colomns (2) and (5) are the corresponding recovered images, and colomns (3) and (6) are the sparse noise terms corresponding to (2) and (5), respectively.

Second, gradient feature is one of the most effective feature extraction methods, and here we directly employ a simple HOG feature for extracting features from scene character images.

Finally, the SRC method (presented in Algorithm 2) is selected for scene character recognition. SRC always can achieve better classification results in comparison with conventional classifiers because the most important nature of SRC is to employ the training samples of all classes to collaboratively represent the test sample. Thus, SRC can sufficiently exploit different classes of samples to represent the testing sample. This is very different from conventional classification methods such as the nearest neighbor (NN) and SVM classifier. The NN classifier only uses one training sample of each class to evaluate the similarity between the training sample and the test sample. The SVM method focuses on training samples near the decision boundary. As a result, SRC can achieve more satisfactory performance in comparison with NN and SVM, and our experimental results in the next section also verify this conclusion.

## IV. EXPERIMENTAL RESULTS

### A. Datasets

We evaluated the proposed method on four publicly available datasets, i.e. Char74K dataset [12], ICADAR 2003 robust character recognition dataset [24], IIIT5K set [25] and Street View Text (SVT) dataset [2]. We only focus on the recognition of English characters and digits, 62 classes in total. Some samples from different datasets have been shown in Fig. 4.

As for the Char74K dataset, we only use the English character images cropped from the natural scene images. It contains 62 character classes. A small subset is used in our experiments, i.e. Char74K-15, which contains 15 training samples per class and 15 test samples per class.

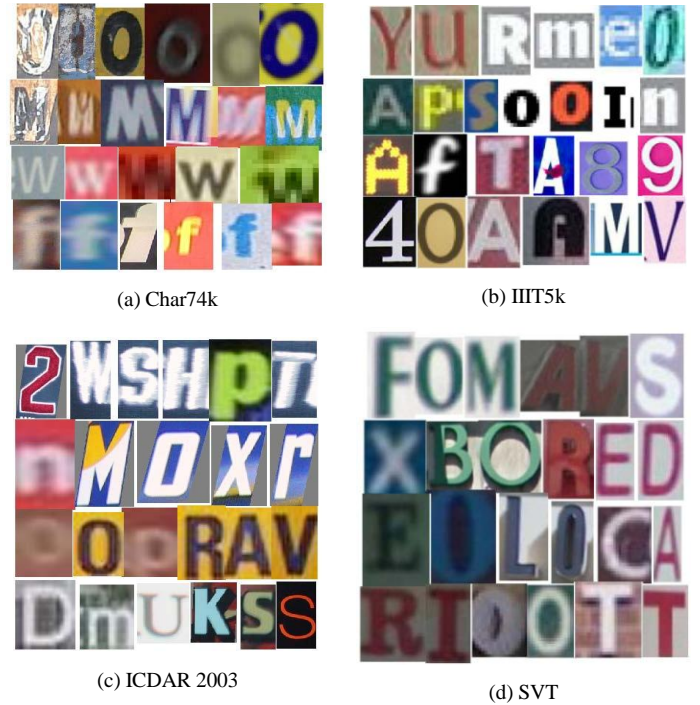

Fig. 4. Sample images of English scene characters from different datasets.

The ICDAR2003 robust reading dataset was collected for the robust reading competition of scene text detection and recognition. The scene character dataset ICDAR03-CH contains more than 11,500 character images. It is a very difficult dataset because of serious non-text background outliers with the cropped character samples, and many character images have very low resolution. There are 6,185 training characters and 5,430 testing characters, respectively. Following the evaluation protocols of literature [6, 11], we excluded some special characters such as '!', and then the final experimental dataset consists of 6,113 characters for training and 5,319 characters for testing.

The IIIT 5K-word (IIIT5K) is composed of 2,000 and 3,000 images for testing and training, respectively. The images contain both scene text images and born-digital images. The character image dataset, which is extracted from word images, is composed of 9,678 samples and 15,269 samples for training and testing, respectively.

The Street View Text (SVT) was collected from Google Street View of road-side scenes. All the images are very challenging because of the large variations in illumination, character image sizes, and variety of font sizes and styles. The SVT character dataset, which was annotated in [17], is utilized for evaluating different scene character recognition methods. This dataset consists of 3,796 character samples from 52 categories (no digit images). SVT character dataset is more difficult to recognize than the ICADAR2003 dataset.

### B. Character Recognition Accuracy

We compare the performance of the proposed method with the recently proposed techniques, including CoHOG [6], ConvCoHOG [11], PHOG [19], MLFP [7], RTPD [26],

GHOG [20], LHOG [20], Kai Wang's methods [2] (HOG+NN, Native Ferns), HOG [17], SBSTR [27], GB [12] (GB+SVM, GB+NN) and convolutional neural network (CNN) [28] on four different datasets. All the images in the experiments are resized into 32×32 pixels and all the images have been first transformed into gray scale images.

For fair comparison, all the experimental protocols are the same as the configurations of literature [6], [11]. In the SVT dataset, only the 3,796 test samples were annotated in [17] for character recognition. Considering that the SVT character dataset has similar distribution to ICDAR2003 and Char74K, we integrated the Char74K EnglishImg dataset and the training samples of ICADAR2003 to construct a new training dataset, which has more than 18,600 characters. We exploited the new training dataset to recognize the test samples of ICDAR2003 and SVT datasets, which is consistent with the experiment setup in [6] [11]. For the datasets of Char74K_15 and IIIT5K, we use their respective training and test datasets for evaluation. Table I shows the experimental results of several recently proposed character recognition methods on four different scene character datasets. For fair comparison, we directly cite all the experimental results from the respective literature. We can see that compare to the best results in the literature, our method can achieve a comparable accuracy on the SVT dataset and a higher accuracy on the char74K_15 dataset. The lowest recognition results of Tesseract OCR technique on different datasets indicate that the conventional OCR techniques can only satisfy scanned document text recognition but not promise challenging scene character recognition. The accuracies of our method on the IIIT5K and ICDAR2003_CH datasets are slightly lower than those of CNN, CoHOG and ConvHOG. This may be attributed to the simple HOG feature used in our method. Moreover, the CNN method was benefitted from a very large number of training samples synthesized in pervious works. We also found that the CNN method removed some bad images and difficult recognized images such that much higher recognition results were achieved. On the other hand, the performance of our method is slightly inferior to CoHOG on the IIIT5K dataset, but on the more difficult SVT dataset, our method performs better.

It is worth noting that our method can achieve the best recognition results in comparison with the HOG feature based recognition methods. For example, the performance of SRC in most cases is better than SVM classifier including linear SVM (Robust PCA+HOG+linear SVM) and nonlinear SVM (Robust PCA+HOG+SVM (RBF kernel)), which were both implemented by ours. Moreover, SRC has obvious superiorities in recognizing scene character images in comparison with the nearest neighbor (NN) classifier. The experimental results also demonstrate that SRC can achieve stable competitive recognition results on different datasets and our method has great potential for scene character recognition.

English scene character recognition is a very difficult task because of no contextual information. There are some tough scene character recognition cases, even for human eyes. For example, distinguishing some upper-case and lower-case characters is very challenging such as 'C' and 'c', 'S' and 's', 'Z' and 'z'. Some letters are very easily confused with numbers, such as letters 'O', 'o' and number '0', letters 'l', 'I' and number '1'.

TABLE I. RECONIGTION RESULTS OF DIFFERENT METHODS ON FOUR SCENE CHARACTER IMAGE DATASETS

| Method | Testing datasets Accuracy | | | |
|---|---|---|---|---|
| | *Char74K_15* | *IIIT5K* | *ICDAR2003_CH* | *SVT* |
| **Robust PCA +HOG+SRC** | **0.67** | 0.76 | 0.79 | **0.75** |
| Robust PCA+ HOG+Linear SVM | 0.63 | 0.74 | 0.75 | 0.73 |
| Robust PCA+ HOG+SVM (RBF) | 0.63 | 0.76 | 0.77 | 0.74 |
| CNN [28] | - | - | **0.84** | - |
| ConvHOG [11] | - | **0.79** | 0.81 | **0.75** |
| CoHOG [6] | - | 0.78 | 0.79 | 0.73 |
| PHOG (Chi-Square Kernel) [19] | - | 0.76 | 0.79 | **0.75** |
| MLFP [7] | 0.64 | - | 0.79 | - |
| RTPD [26] | - | - | 0.76 | 0.67 |
| GHOG+SVM [20] | 0.62 | - | 0.76 | - |
| LHOG+SVM [20] | 0.58 | - | 0.75 | - |
| SBSTR [27] | 0.60 | 0.63 | - | - |
| HOG+NN [2] | 0.58 | 0.68 | 0.52 | 0.68 |
| NATIVE+FERNS [2] | 0.54 | - | 0.64 | - |
| GB+SVM [12] | 0.53 | - | - | - |
| GB+NN [12] | 0.47 | - | - | - |
| Tesseract OCR [3] | - | 0.32 | 0.37 | 0.35 |

Character image segmentation results also can greatly influence the performance of recognition results. Fig. 5 and Fig. 6 respectively show some examples of correctly and incorrectly recognized character images. We can see that more accurate segmentation has higher potential of correct recognition. Moreover, serious skewed images are more difficult to be exactly recognized in comparison with that of well-aligned images and many mis-recognized examples are usually low-resolution, ambiguous and distorted images.

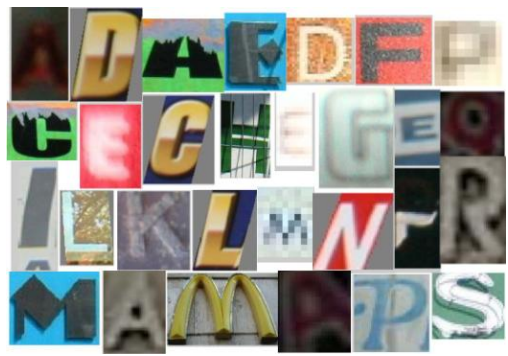

Fig. 5. Examples of correctly recognized images.

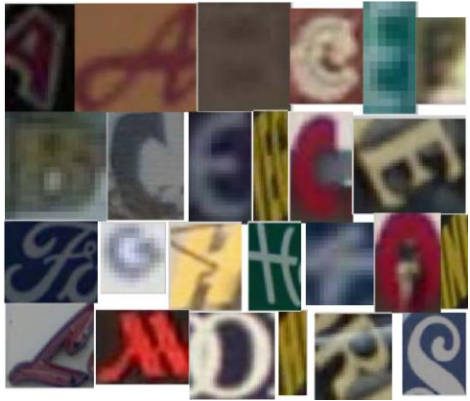

Fig. 6. Examples of incorrectly recognized images.

## V. CONCLUSION AND FUTURE WORK

We proposed a novel robust recognition method for natural scene characters by exploiting the robust principle component analysis, HOG feature extraction and sparse representation based classification. The robustness is mainly due to the low-rank image recovery and sparse representation. Experimental results on four different scene character datasets demonstrate the robustness and effectiveness of our method, and the performance of our method is competitive with the current the-state-of-the-art algorithms.

Robust PCA method assumes that the underlying data structure is from a single low-rank space, but real data is always drawn from a union of multiple subspaces. Thus, one of our future works is to develop a more effective low-rank representation based image denoising method. As for image feature extraction, we only use the most simple HOG feature in our experiments. We believe the recognition performance can be further improved by using more discriminative features.

## VI. ACKNOWLEDGEMENTS

This work has been supported in part by the National Basic Research Program of China (973 Program) Grant 2012CB316302, the National Natural Science Foundation of China (NSFC) Grant 61411136002, the Strategic Priority Research Program of the Chinese Academy of Sciences (Grant XDA06040102), and the Xinjiang Uygur Autonomous Region Science and Technology Project (Grant 201230122).

## REFERENCES

[1] Q. Ye, D. Doermann, "Text detection and recognition in imagery: A survey," *IEEE Trans. Pattern Anal. Mach. Intell.*, vol.37(7), pp. 1480-1500, 2015.

[2] K. Wang, B. Babenko, S. Belongie, "End-to-end scene text recognition," *in Proceeding of 2011 ICCV*, Barcelona, Spain, pp. 1457-1464, 2011.

[3] Google Tesseract OCR, http://code.google.com/p/tesseract-ocr/.

[4] M. K. Zhou, X. Y. Zhang, F. Yin and C. L. Liu, "Discriminative quadratic feature learning for handwritten Chinese character recognition," *Pattern Recognition*, vol. 49, pp. 7–18, 2016.

[5] U. Roy, A. Mishra, K. Alahari, C.V. Jawahar, "Scene Text Recognition and Retrieval for Large Lexicons," *in Proceeding of ACCV*, Singapore, pp. 494-508, 2014.

[6] S. Tian, S. Lu, B. Su, C.L. Tan, "Scene text recognition using co-occurrence of histogram of oriented gradients," *in Proceeding of 12th ICDAR*, Washington, D.C., USA, pp. 912-916, 2013.

[7] C. Y. Lee, A. Bhardwaj, W. Di, V. Jagadeesh, R. Piramuthu, "Region-based discriminative feature pooling for scene text recognition," *in Proceeding of 2014 CVPR*, Columbus, OH, USA, pp. 4050-4057, 2014.

[8] X. Chen, A. L. Yuille, "Detecting and reading text in natural scenes," *in Proceeding of 2004 CVPR*, Washington, DC, USA, vol. 2, pp. II-366-II-373, 2004.

[9] K. Kita, T. Wakahara, "Binarization of color characters in scene images using k-means clustering and support vector machines," *in Proceeding of 20th ICPR*, Istanbul, Turkey, pp. 3183-3186, 2010.

[10] A. Bissacco, M. Cummins, Y. Netzer, H. Neven, "PhotoOCR: Reading Text in Uncontrolled Conditions," *in Proceeding of 14th ICCV*, Sydney, Australia, pp.785-792, 2013.

[11] B. Su, S. Lu, S. Tian, J. H. Lim, C. L. Tan, "Character Recognition in Natural Scenes using Convolutional Co-occurrence HOG," *in Proceeding of 22nd ICPR*, Istanbul, Turkey, pp. 2926-2931, 2014.

[12] T. E. de Campos, B. R. Babu, M. Varma, "Character Recognition in Natural Images," *in Proceeding of VISAPP*, Lisboa, Portugal, vol. 2, pp. 273-280, 2009.

[13] E. J. Candès, X. Li, Y. Ma, J. Wright, "Robust principal component analysis?" *Journal of the ACM*, vol. 58(3), pp. 11, 2011.

[14] J. Wright, A.Y. Yang, A. Ganesh, S. S. Sastry, Y. Ma, "Robust face recognition via sparse representation," *IEEE Trans. Pattern Anal. Mach. Intel.*, vol. 31(2), pp. 210-227, 2009.

[15] T. Zhou, D. Tao, "Godec: Randomized low-rank & sparse matrix decomposition in noisy case," *in Proceeding of 28th ICML*, Bellevue, USA, pp. 33 - 40, 2011.

[16] C. F. Chen, C. P. Wei, Y. Wang, "Low-rank matrix recovery with structural incoherence for robust face recognition," *in Proceeding of 2012 CVPR*, Providence, USA, pp. 2618-2625, 2012.

[17] A. Mishra, K. Alahari, C.V. Jawahar, "Top-down and bottom-up cues for scene text recognition," *in Proceeding of 2012 CVPR,* Providence, USA, pp. 2687-2694, 2012.

[18] K. Wang, S. Belongie, "Word spotting in the wild," *in Proceeding of ECCV*, Heraklion, Greece, pp. 591-604, 2010.

[19] Z. R. Tan, S. Tian and C. L. Tan, "Using pyramid of histogram of oriented gradients on natural scene text recognition," *in Proceeding of 2014 ICIP,* Paris, France, pp. 2629-2633, 2014.

[20] C. Yi, X. Yang and Y. Tian, "Feature representations for scene text character recognition: A comparative study," *in Proceeding of 12th ICDAR*, Washington, DC, USA, pp. 907-911, 2013.

[21] D. Wang, H. Wang, D. Zhang, J. Li, D. Zhang, Robust scene text recognition using sparse coding based faetures, arXiv:1512.08669

[22] Z. Zhang, Y. Xu, J. Yang, X. Li, D. Zhang, "A survey of sparse representation: algorithms and applications," *IEEE Access*, vol. 3, pp. 490-530, 2015.

[23] Z. Lin, M. Chen and Y. Ma, "The augmented lagrange multiplier method for exact recovery of corrupted low-rank matrices," arXiv preprint arXiv:1009.5055, 2010.

[24] S. M. Lucas, A. Panaretos, L. Sosa, A. Tamg, S. Wong and R. Young, "ICDAR 2003 robust reading competitions," *in Proceeding of 7th ICDAR*, pp. 682-687, 2003.

[25] A. Mishra, K. Alahari, C. V. Jawahar, "Scene text recognition using higher order language priors," *in Proceeding of 23rd BMVC*, Surrey, United Kingdom, 2012.

[26] T. Q. Phan, P. Shivakumara, S. Tian, C. L. Tan, "Recognizing text with perspective distortion in natural scenes," *in Proceeding of 14th ICCV*, Sydney, Australia, pp. 569-576, 2013.

[27] C. Yi and Y. Tian, "Scene text recognition in mobile applications by character descriptor and structure configuration," *IEEE Trans. Image Processing*, vol. 23(7), pp. 2972-2982, 2014.

[28] T. Wang, D. J. Wu, A. Coates A, A.Y. Ng, "End-to-end text recognition with convolutional neural networks," *in Proceeding of 21st ICPR*, Tsukuba, pp. 3304-3308, 2012.